\documentclass{article}
\usepackage{pifont}
\usepackage{amsfonts}
\usepackage{tabularx}
\usepackage{graphicx} % 用于插入图片
\usepackage{float}    % 用于使用[H]等位置选项
\usepackage{pifont}
\usepackage{booktabs}
\usepackage{spconf,amsmath,graphicx,hyperref,booktabs}
\usepackage{multirow}
\usepackage{url}
\usepackage{hyperref}

\title{CONQUER: CONtext-aware representation with QUery Enhancement for Text-Based Person Search}
\name{Zequn Xie
  \thanks{$^{\dagger}$ These authors contributed equally to this work.}
  \thanks{$^{*}$ Corresponding author. Email: hanxiaosong@jlu.edu.cn}
}

\address{Zhejiang University, Hangzhou, China }

\begin{document}
\ninept
\maketitle
\begin{abstract}
Text-Based Person Search (TBPS) aims to retrieve pedestrian images from large galleries using natural language descriptions. This task, essential for public safety applications, is hindered by cross-modal discrepancies and ambiguous user queries. We introduce CONQUER, a two-stage framework designed to address these challenges by enhancing cross-modal alignment during training and adaptively refining queries at inference. During training, CONQUER employs multi-granularity encoding, complementary pair mining, and context-guided optimal matching based on Optimal Transport to learn robust embeddings. At inference, a plug-and-play query enhancement module refines vague or incomplete queries via anchor selection and attribute-driven enrichment, without requiring retraining of the backbone. Extensive experiments on CUHK-PEDES, ICFG-PEDES, and RSTPReid demonstrate that CONQUER consistently outperforms strong baselines in both Rank-1 accuracy and mAP, yielding notable improvements in cross-domain and incomplete-query scenarios. These results highlight CONQUER as a practical and effective solution for real-world TBPS deployment. Source code is available at \url{https://github.com/zqxie77/CONQUER}.

\end{abstract}

\section{Introduction}
\label{sec:intro}

Text-based Person Search (TBPS) is a cross-modal retrieval task that locates target individuals in large image galleries using natural language descriptions~\cite{1}. As a bridge between Person Re-identification (Re-ID) and text-to-image retrieval, TBPS allows for more flexible and descriptive queries than traditional image-based methods. However, TBPS remains challenging due to redundant visual features~\cite{2,3} and the inherent noise in cross-modal correspondence~\cite{xie2025dynamic}. Furthermore, the lack of adaptive mechanisms for handling ambiguous queries during inference limits practical application.

\begin{figure}[h!]
\centering
\includegraphics[width=\linewidth]{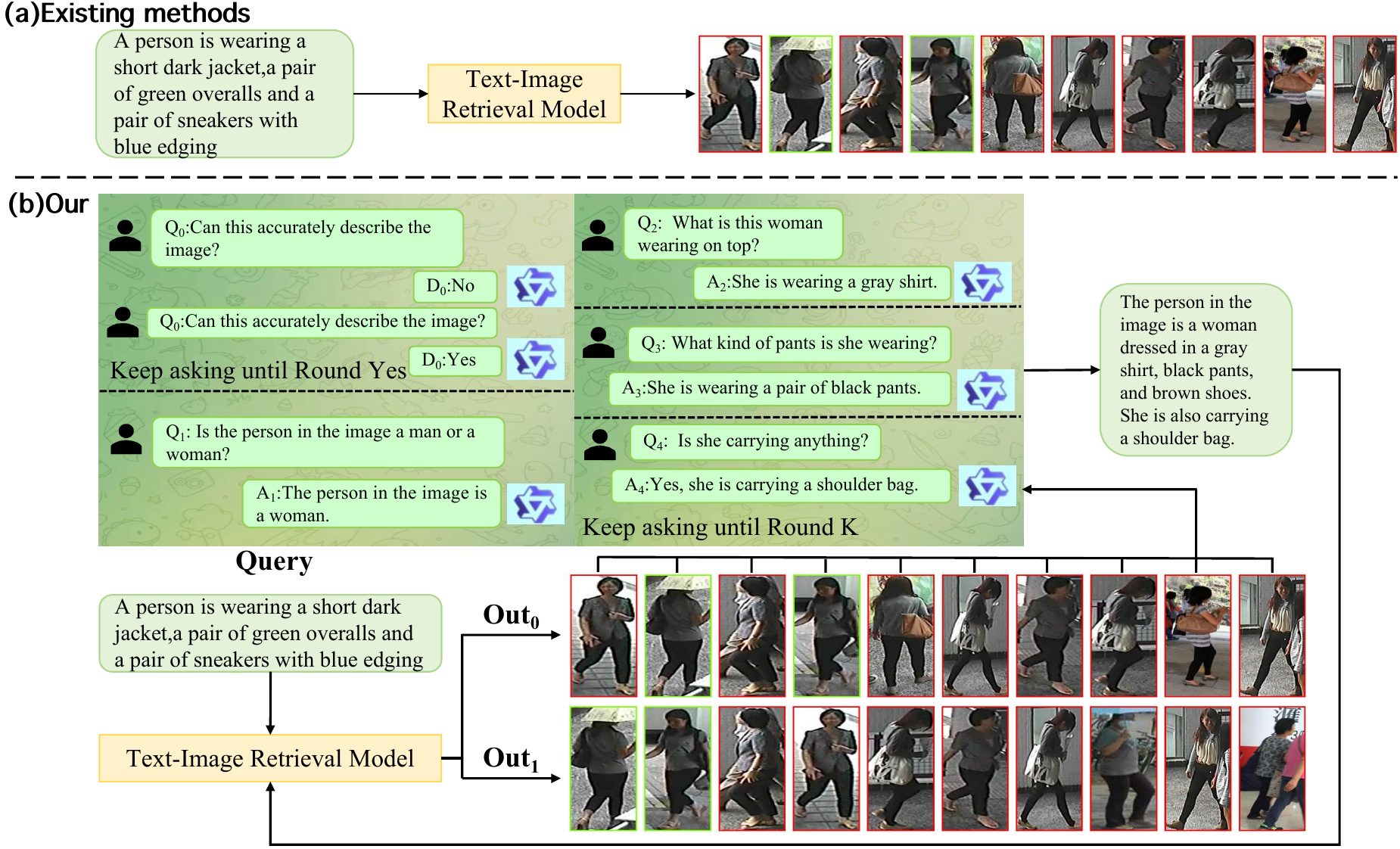}
\caption{A comparison of TBPS.(a) \textbf{Existing Method:} The search is performed directly using the original text query.(b) \textbf{Our IQE Approach:} We improve the query at inference time without retraining. First, our method finds a relevant anchor image. Then, an MLLM learns key visual details from this image through a Q\&A process. Finally, it fuses these details with the original text to create an improved query and re-ranks the search results.}
\label{fig:iqe}
\end{figure}

Most existing methods improve cross-modal alignment using fine-grained annotations or large pre-trained models like CLIP~\cite{4}. While integrated planning and scheduling frameworks have proven effective in optimizing complex systems~\cite{liu2025integratedplanningmachinelevelscheduling}, TBPS methods still mainly rely on \textit{passive alignment}—matching features in the embedding space without actively improving the inputs. Some approaches explore hierarchical visual perception to achieve deeper alignment~\cite{xie2026delvingdeeperhierarchicalvisual,xie2026hvdhumanvisiondrivenvideo}, yet they also suffer from this passive limitation. This leads to two key problems: (1) Methods such as CFAM~\cite{5} depend on manual annotations, but their performance drops when annotations are scarce; (2) Methods like DM-Adapter~\cite{6} and MGRL~\cite{7} cannot refine brief or ambiguous queries at inference, resulting in poor results for incomplete descriptions.

Some recent works have attempted active refinement or interaction. For instance, chat-driven mechanisms have been proposed to generate text for retrieval~\cite{xie2025chat}, though purely generative approaches may hallucinate attributes. Other works like TVFR~\cite{8} are sensitive to the initial query, and AUL~\cite{9} mainly helps during training. \textit{These limitations motivate us to design a method that achieves fine-grained cross-modal alignment during training and actively refines ambiguous or incomplete queries at inference.}

To solve the aforementioned problems, we propose \textbf{CONQUER}, a two-stage framework as illustrated in Fig.~\ref{fig:iqe}. During training, our \textbf{Context-Aware Representation Enhancement (CARE)} module learns robust cross-modal embeddings. Subsequently, at the inference stage, our \textbf{Interactive Query Enhancement (IQE)} module acts as a plug-and-play enhancement that adaptively refines user queries. By leveraging fine-grained attributes extracted from high-confidence candidate images, IQE enriches ambiguous or underspecified descriptions. The enhanced information is then fused with the original query, enabling more accurate re-ranking of the gallery and improved retrieval results.

Our main contributions are as follows:
\begin{itemize}
    \item We introduce CARE, a context-aware representation enhancement module that addresses the cross-modal gap by combining multi-granularity representation encoding, complementary pair mining, and context-guided optimal matching via Optimal Transport.
    \item We develop IQE, an interactive query enhancement strategy that adaptively refines ambiguous or incomplete user queries during inference by leveraging attribute information from candidate images.
    
    \item Extensive experiments show that CONQUER achieves state-of-the-art performance on several TBPS benchmarks and demonstrates strong robustness in cross-domain and incomplete-query scenarios.
\end{itemize}

\section{Methodology}

Context-aware Representation with Query Enhancement for Retrieval (CONQUER) is a two-stage framework for TBPS that comprises CARE during training and IQE at inference. CARE enhances cross-modal embeddings by leveraging contextual cues, while IQE adaptively refines user queries to improve retrieval accuracy.

\begin{figure*}[!t]
    \centering
    % NOTE: Ensure "figure2.png" is in the same folder as.tex file
    \includegraphics[width=1\textwidth, height=0.4\textheight]{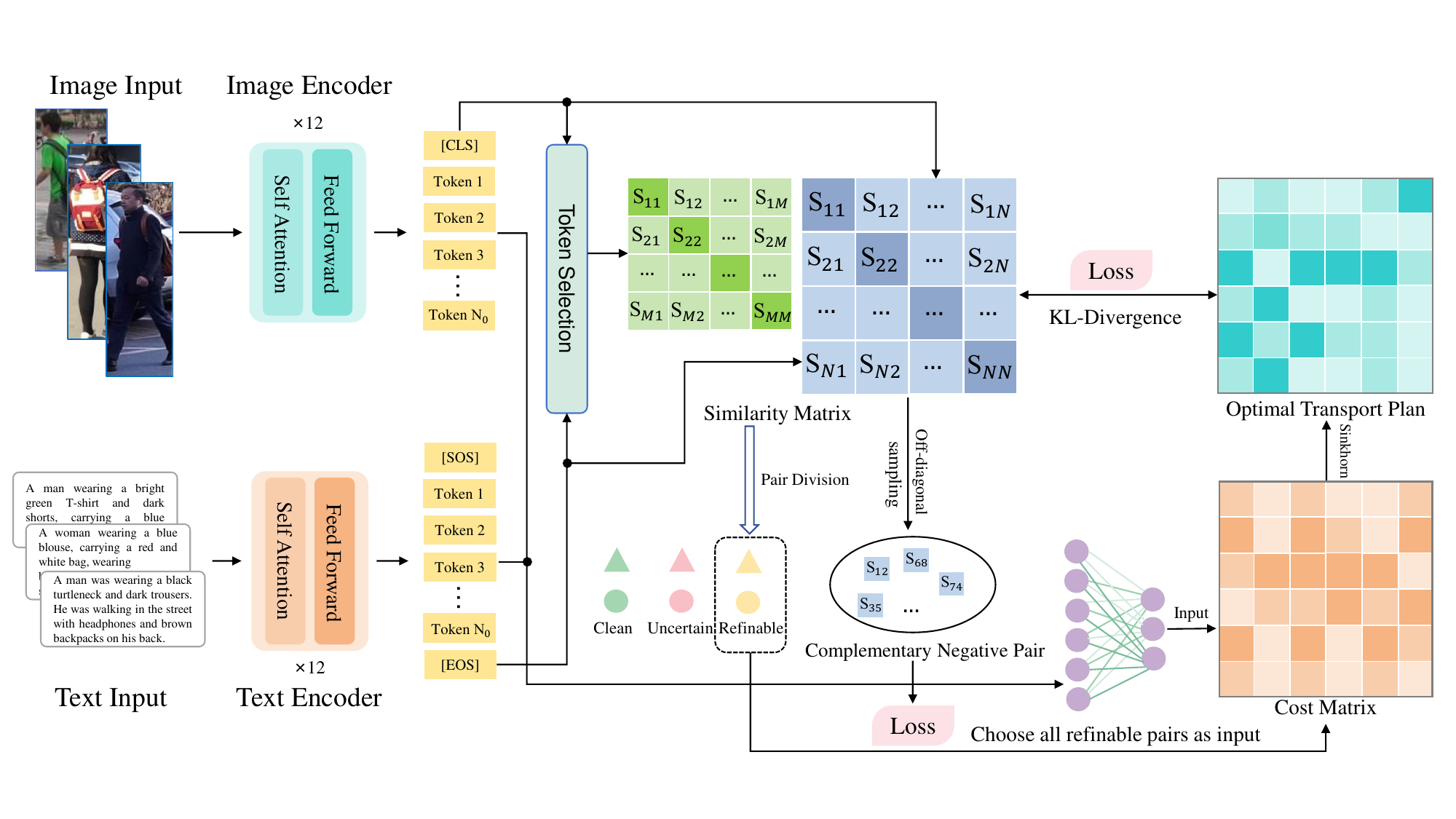}  
   \caption{\textbf{The Architecture of the Context-Aware Representation Enhancement (CARE) Module.} The CARE module leverages multi-granularity representation encoding to learn robust cross-modal alignments. For complementary pair mining, it first classifies training pairs into clean, uncertain, and refinable sets by jointly analyzing similarity matrices from both global features and selected local tokens; hard negatives are then mined from the identified `refinable' set. Simultaneously, its context-guided optimal matching component employs an Optimal Transport (OT) solver to align fine-grained features. This local alignment is in turn guided by the global similarities via a KL-Divergence loss, ensuring feature matching remains consistent with the overall semantic context.} 
    \label{fig:1}
\end{figure*}

\subsection{Context-Aware Representation Enhancement}
As illustrated in Fig.~\ref{fig:1}, CARE improves cross-modal alignment through three main components: multi-granularity representation encoding, complementary pair mining, and context-guided optimal matching.

\subsubsection{Multi-granularity Representation Encoding}
Let $\mathbf{I}$ and $\mathbf{T}$ denote the input image and text. Modality-specific encoders $E_I(\cdot)$ and $E_T(\cdot)$ extract both global and local features. For each image-text pair, local features are $\{\mathbf{v}_m\}_{m=1}^M$ (visual tokens) and $\{\mathbf{w}_n\}_{n=1}^N$ (text tokens), with global features $\bar{\mathbf{v}}_i$ and $\bar{\mathbf{w}}_i$ for the $i$-th sample.

Arrange local features into matrices:
\begin{equation}
    \mathbf{V} = [\mathbf{v}_1, \dots, \mathbf{v}_M] \in \mathbb{R}^{d \times M}, \quad
    \mathbf{W} = [\mathbf{w}_1, \dots, \mathbf{w}_N] \in \mathbb{R}^{d \times N}.
\end{equation}

The batch-wise cross-modal similarity matrix $S \in \mathbb{R}^{B \times B}$ is computed from global features:
\begin{equation}
    S_{ij} = \frac{\bar{\mathbf{v}}_i^\top \bar{\mathbf{w}}_j}{\|\bar{\mathbf{v}}_i\|\|\bar{\mathbf{w}}_j\|},
\end{equation}
where $B$ is the batch size and $S_{ij}$ is the cosine similarity between the $i$-th image and $j$-th text.

Based on confidence, training pairs are divided into high-confidence (clean), low-confidence (uncertain), and refinable sets.

\subsubsection{Complementary Pair Mining}
For refinable pairs, off-diagonal sampling generates complementary negatives:
\begin{equation}
    \mathcal{N} = \{ S_{ij} \mid i \neq j, \, (i,j) \in \text{Refinable} \},
\end{equation}
enhancing negative diversity and sharpening decision boundaries.

\subsubsection{Context-guided Optimal Matching}
Cross-modal alignment refinement is formulated as an Optimal Transport (OT) problem.
Given local features $\{\mathbf{v}_m\}_{m=1}^M$ and $\{\mathbf{w}_n\}_{n=1}^N$, a learnable cost matrix is computed:
\begin{equation}
    C_{mn} = f_\theta(\mathbf{v}_m, \mathbf{w}_n),
\end{equation}
where $f_\theta(\cdot)$ is a neural network estimating contextual dissimilarity.

The OT objective seeks a transport plan $T \in \mathbb{R}^{M \times N}$ by minimizing:
\begin{equation}
    \min_{T \in \Pi(\mathbf{a}, \mathbf{b})} \langle T, C \rangle + \lambda \, \mathrm{H}(T),
\end{equation}
where $\Pi(\mathbf{a},\mathbf{b})$ is the set of valid transport plans. We solve this regularized OT problem using the Sinkhorn algorithm.

For supervision, compute a local similarity matrix:
\begin{equation}
    S^{\mathrm{loc}}_{mn} = \frac{\mathbf{v}_m^\top \mathbf{w}_n}{\|\mathbf{v}_m\|\|\mathbf{w}_n\|}.
\end{equation}

Align $T$ with $S^{\mathrm{loc}}$ by minimizing their row-wise KL-divergence:
\begin{equation}
    \mathcal{L}_{\mathrm{OT}} = \mathrm{KL}(\mathrm{softmax}_{\mathrm{row}}(T) \; \| \; \mathrm{softmax}_{\mathrm{row}}(S^{\mathrm{loc}})).
\end{equation}

The final CARE loss is:
\begin{equation}\label{eq:care_loss}
    \mathcal{L}_{\mathrm{CARE}} = \mathcal{L}_{\mathrm{align}} + \alpha \mathcal{L}_{\mathrm{neg}} + \beta \mathcal{L}_{\mathrm{OT}},
\end{equation}
where $\mathcal{L}_{\mathrm{align}}$ is cross-modal alignment loss, $\mathcal{L}_{\mathrm{neg}}$ is the loss for complementary negatives, and $\alpha, \beta$ are trade-off weights.

\subsection{Interactive Query Enhancement }
As illustrated in Fig.~\ref{fig:iqe}, IQE is a plug-and-play inference module that refines ambiguous queries without retraining the backbone. The process consists of three main steps: anchor identification, interactive query refinement, and query fusion for re-ranking.

% continue numbering from Section 2.1 (last eq. (8))
\setcounter{equation}{8}

\subsubsection{Anchor Identification}
Given a query $T$, the backbone returns top-$K$ candidates $\{\hat I_j\}_{j=1}^K$ with similarity scores $\{s_j\}$. A multimodal reasoning model produces verification confidences $v_j$, and we define the anchor set as:
\begin{equation}
A=\{\hat I_j \mid v_j\ge\psi\}.
\end{equation}
To control cost, IQE employs early-stop (accept initial ranking if $s_1>\xi$) and a fallback trigger (activate IQE only when the query is short/ambiguous or retrieval confidence is low).

\subsubsection{Interactive Query Refinement}
For each anchor $a\in A$, diagnostic questions $C_a=\mathcal{G}_{\text{ques}}(T,a)$ are generated, with the detailed strategy provided in our code repository. Answers from the MLLM are collected as:
\begin{equation}
r_a = \{(c_i,\, a_i,\, p_{i})\}_{i=1}^{|C_a|},
\end{equation}
and only high-confidence responses ($p_i\ge\tau$) are retained. Answers across anchors are aggregated via confidence-weighted voting:
\[
\mathrm{score}(u)=\frac{1}{|A|}\sum_{a\in A}\sum_{(c,a',p)\in r_a}\mathbf{1}[a'\equiv u]\cdot p,
\]
and the evidence set is $R=\{u\mid \mathrm{score}(u)\ge\eta\}$. Finally, the reconstructor synthesizes the enhanced query:
\begin{equation}
T'=\mathrm{MLLM}_{\text{agg}}(T,R).
\end{equation}

\subsubsection{Query Fusion and Re-ranking}
The final retrieval score combines original and enhanced similarities, with an optional anchor bonus:
\begin{equation}
\mathrm{Score}(I)=\gamma\cdot\mathrm{sim}(T,I)+(1-\gamma)\cdot\mathrm{sim}(T',I)+\beta\cdot\mathbf{1}[I\in A].
\end{equation}
A safeguard reverts to the original ranking if $T'$ causes a substantial drop in validation alignment. Further implementation details will be released with the code.

\section{Experiments}

\subsection{Experimental Settings}
\noindent\textbf{Datasets and Metrics.}
We evaluate CONQUER on three standard TBPS benchmarks: CUHK-PEDES, ICFG-PEDES, and RSTPReid, following their official train/validation/test splits. 
Performance is measured using Rank-$k$ accuracy ($k=1,5$) and mean Average Precision (mAP) under the text-to-image retrieval setting.

\noindent\textbf{Implementation Details.}
The visual and textual backbones are initialized with CLIP ViT-B/16. 
The CARE module is trained using the Adam optimizer (batch size 64, learning rate $1\times10^{-4}$ with cosine decay) for 60 epochs. 
The loss weights in Eq.~\ref{eq:care_loss} are set to $\alpha=0.5$ and $\beta=0.1$. 
For IQE, we employ Qwen2.5-VL-7B as the reasoning engine, fine-tuned via LoRA. 
Inference hyperparameters are set as: top-$K$ anchors $K=5$, re-ranking weight $\gamma=0.6$, and thresholds $\psi=0.90$, $\xi=0.85$, $\tau=0.85$, and $\eta=0.5$.
\begin{table*}[!t]
\centering
\caption{Performance comparison with state-of-the-art methods on CUHK-PEDES, ICFG-PEDES, and RSTPReid datasets.
The best results are in \textbf{bold} and the second best are \underline{underlined}.}
\label{tab:sota_comparison}
\resizebox{\textwidth}{!}{%
\begin{tabular}{l|ccc|ccc|ccc|ccc}
\toprule
\multirow{2}{*}{\textbf{Methods}} & \multirow{2}{*}{\textbf{Ref.}} & \multirow{2}{*}{\textbf{Image Enc.}} & \multirow{2}{*}{\textbf{Text Enc.}}
& \multicolumn{3}{c|}{\textbf{CUHK-PEDES}}
& \multicolumn{3}{c|}{\textbf{ICFG-PEDES}}
& \multicolumn{3}{c}{\textbf{RSTPReid}} \\
\cmidrule(lr){5-7} \cmidrule(lr){8-10} \cmidrule(lr){11-13}
& & & & \textbf{R@1} & \textbf{R@5} & \textbf{mAP}
& \textbf{R@1} & \textbf{R@5} & \textbf{mAP}
& \textbf{R@1} & \textbf{R@5} & \textbf{mAP} \\
\midrule
ViTAA \cite{10} & ECCV20 & RN50 & LSTM & 54.92 & 75.18 & 51.60 & 50.98 & 68.79 & - & - & - & - \\
SSAN \cite{11} & arXiv21 & RN50 & LSTM & 61.37 & 80.15 & - & 54.23 & 72.63 & - & 43.50 & 67.80 & - \\
IVT \cite{12} & ECCV22 & ViT-Base & BERT & 65.59 & 83.11 & 60.66 & 56.04 & 73.60 & - & 46.70 & 70.00 & - \\
BEAT \cite{13} & MM23 & RN101 & BERT & 65.61 & 83.45 & - & 58.25 & 75.92 & - & 48.10 & 73.10 & - \\
LCR$^2$S \cite{14} & MM23 & RN50 & TextCNN & 67.36 & 84.19 & 59.24 & 57.93 & 76.08 & 38.21 & 54.95 & 76.65 & 40.92 \\
UniPT \cite{15} & ICCV23 & CLIP-ViT & CLIP-Xformer & 68.50 & 84.67 & - & 60.09 & 76.19 & - & 51.85 & 74.85 & - \\
CFine \cite{16} & TIP23 & CLIP-ViT & BERT & 69.57 & 85.93 & - & 60.83 & 76.55 & - & 50.55 & 72.50 & - \\
DM-Adapter \cite{6} & AAAI25 & CLIP-ViT & CLIP-Xformer & 72.17 & 88.74 & 64.33 & 62.64 & 79.53 & 36.50 & 60.00 & 82.10 & 47.37 \\
IRRA \cite{3} & CVPR23 & CLIP-ViT & CLIP-Xformer & 73.38 & 89.93 & 66.13 & 63.46 & 80.25 & 38.06 & 60.20 & 81.30 & 47.17 \\
TBPS \cite{17} & AAAI24 & CLIP-ViT & CLIP-Xformer & 73.54 & 88.19 & 65.38 & 65.05 & 80.34 & 39.83 & 61.95 & 83.55 & 48.26 \\
MGRL \cite{7} & ICASSP24 & CLIP-ViT & CLIP-Xformer & 73.91 & \underline{90.68} & 67.28 & 63.87 & \underline{82.34} & 39.12 & - & - & - \\

DCEL \cite{19} & MM23 & CLIP-ViT & CLIP-Xformer & 75.02 & \textbf{90.89} & - & 64.88 & 81.34 & - & 61.35 & {83.95} & - \\
OCDL \cite{20} & ICASSP25 & CLIP-ViT & CLIP-Xformer & 75.10 & 89.43 & \underline{68.18} & 64.53 & 80.23 & \textbf{40.76} & 61.60 & 82.35 & 49.77 \\
CFAM \cite{5} & CVPR24 & CLIP-ViT & CLIP-Xformer & 75.60 & 90.53 & 67.27 & 65.38 & 81.17 & 39.42 & 62.45 & 83.55 & 49.50 \\
DP \cite{21} & AAAI24 & CLIP-ViT & CLIP-Xformer & 75.66 & 90.59 & 66.58 & 65.61 & 81.73 & 39.14 & 62.48 & 83.77 & 48.86 \\
RDE \cite{22} & CVPR24 & CLIP-ViT & CLIP-Xformer & \underline{75.94} & 90.14 & 67.56 & \underline{67.68} & \textbf{82.47} & 40.06 & {65.35} & 83.95& {50.88} \\
CTGI \cite{xie2025chat} & EMNLP25 & CLIP-ViT & CLIP-Xformer & 67.82 & 85.45 &  55.14 & 56.16& 73.18 & 32.40 & \underline{66.35} & \textbf{85.50}& \underline{51.51} \\
\hline
\textbf{CONQUER (Ours)} & - & CLIP-ViT & CLIP-Xformer & \textbf{77.13} & 90.06 & \textbf{68.75} & \textbf{67.70} & 81.88 & \underline{40.36} & \textbf{68.40} & \underline{84.95} & \textbf{51.73} \\
\bottomrule
\end{tabular}%
}
\end{table*}

\subsection{ Comparisons with State-of-the-Art Methods}

In this subsection, as shown in Table \ref{tab:sota_comparison}, we present a comparison of CONQUER with existing methods on three TBPS datasets. On the CUHK-PEDES dataset, with CLIP-ViT as the backbone, CONQUER achieves the best results, obtaining 77.13\% R@1 and 68.75\% mAP, and outperforms recent methods such as OCDL \cite{20} and RDE \cite{22}. For the ICFG-PEDES dataset, it also delivers strong performance with CLIP-ViT, reaching 67.70\% R@1 and 40.36\% mAP. Since it surpasses methods including IRRA \cite{3} and DCEL \cite{19}, our method remains a leading solution with superior overall performance across multiple datasets. On the RSTPReid dataset, our approach further performs excellently, outperforming RDE \cite{22} with 68.40\% R@1 and 51.73\% mAP, and ranks the highest among all methods. In conclusion, our method consistently outperforms most state-of-the-art techniques across all three datasets, thus demonstrating its superior performance and robustness.

\subsection{Robustness of the Method}

To further assess the generalization ability of CONQUER, we conduct comprehensive cross-domain experiments on widely used TBPS benchmarks. As summarized in Table~\ref{tab:2}, our approach consistently achieves superior results compared to representative methods under all transfer settings. In particular, CONQUER surpasses strong baselines such as IRRA and SEN not only in R@1 but also in R@5 and mAP, indicating stable gains across different evaluation criteria. The transfer from CUHK-PEDES to RSTPReid and from RSTPReid to CUHK-PEDES is especially challenging due to large domain gaps, yet CONQUER improves R@1 by 3.10\% and 7.87\% over SEN, respectively. These results highlight that our CARE and IQE modules jointly enhance the robustness of the learned representations, enabling effective adaptation to unseen domains and making the framework highly suitable for real-world deployment.

\vspace{-1em}
\begin{table}[H]
\caption{Cross-domain performance evaluation. The notation ``Source $\rightarrow$ Target'' denotes that the model is trained on the source dataset and tested on the target dataset. Abbreviations used: CUHK (CUHK-PEDES), ICFG (ICFG-PEDES), and RSTP (RSTPReid). The best results are highlighted in \textbf{bold}, and the second-best are \underline{underlined}.}
\label{tab:2}
\centering
\scalebox{0.9}{
\begin{tabular}{c|c|ccc}
\toprule
\textbf{Method} & \textbf{Domain} & \textbf{R@1} & \textbf{R@5} & \textbf{mAP} \\
\midrule
\multirow{6}{*}{{IRRA} ~\cite{3}}
 & CUHK $\rightarrow$ RSTP & 53.30 & \underline{77.15} & 39.63 \\
 & CUHK $\rightarrow$ ICFG & 42.42 & 62.11 & 21.77 \\
 & ICFG $\rightarrow$ RSTP & 45.30 & 69.35 & 36.83 \\
 & ICFG $\rightarrow$ CUHK & 33.46 & 56.30 & 31.56 \\
 & RSTP $\rightarrow$ ICFG & 32.30 & 49.69 & 20.54 \\
 & RSTP $\rightarrow$ CUHK & 32.80 & 55.25 & 30.29 \\
\midrule
\multirow{6}{*}{{SEN} ~\cite{23}}
 & CUHK $\rightarrow$ RSTP & \underline{55.50} & \textbf{77.85} & \textbf{45.29} \\
 & CUHK $\rightarrow$ ICFG & \underline{45.34} & \underline{63.45} & \underline{23.26} \\
 & ICFG $\rightarrow$ RSTP & \underline{47.45} & \textbf{71.95} & \textbf{39.86} \\
 & ICFG $\rightarrow$ CUHK & \underline{37.88} & \textbf{60.48} & \underline{35.07} \\
 & RSTP $\rightarrow$ ICFG & \underline{36.23} & \underline{53.31} & \underline{22.32} \\
 & RSTP $\rightarrow$ CUHK & \underline{35.40} & \textbf{57.71} & \underline{33.41} \\
\midrule
\multirow{6}{*}{\textbf{CONQUER (Ours)}}
 & CUHK $\rightarrow$ RSTP & \textbf{58.60} & 77.00 & \underline{42.20} \\
 & CUHK $\rightarrow$ ICFG & \textbf{48.59} & \textbf{65.43} & \textbf{24.65} \\
 & ICFG $\rightarrow$ RSTP & \textbf{54.55} & \underline{71.60} & \underline{39.44} \\
 & ICFG $\rightarrow$ CUHK & \textbf{40.98} & \underline{60.33} & \textbf{36.18} \\
 & RSTP $\rightarrow$ ICFG & \textbf{44.81} & \textbf{56.60} & \textbf{26.58} \\
 & RSTP $\rightarrow$ CUHK & \textbf{43.27} & \underline{57.37} & \textbf{35.08} \\
\bottomrule
\end{tabular}
}
\end{table}
\vspace{-1em}

\subsection{Ablation Studies}

To verify the specific contributions of the CARE and IQE modules, we conducted comprehensive ablation studies on the RSTPReid dataset, as detailed in Table~\ref{tab:ablation}. The baseline model serves as a starting point, achieving a Rank-1 accuracy of 66.50\% and an mAP of 51.47\%. When the CARE or IQE modules are deployed individually, we observe only marginal improvements over the baseline. This indicates that neither module alone is sufficient to fully address the cross-modal challenges. However, a significant performance boost is realized when both modules are integrated, raising the Rank-1 to 68.40\% and mAP to 51.73\%. This result clearly demonstrates the synergistic effect between the two components: CARE optimizes feature alignment during the training phase, while IQE refines the retrieval targets during inference. Together, they effectively complement each other to achieve the optimal experimental results.

\begin{table}[t]
\centering
\caption{Ablation study of CONQUER components on RSTPReid.}
\label{tab:ablation}
\resizebox{\columnwidth}{!}{%
\begin{tabular}{ccccc c ccc}
\toprule
\multirow{2}{*}{\textbf{Index}} & \multirow{2}{*}{\textbf{Base}} & \multicolumn{2}{c}{\textbf{Components}} & \phantom{a} & \multicolumn{3}{c}{\textbf{Retrieval Performance}} \\
\cmidrule{3-4} \cmidrule{6-8}
& & \textbf{CARE} & \textbf{IQE} && \textbf{R@1} & \textbf{R@5} & \textbf{mAP} \\
\midrule
1 & \checkmark   &   & \checkmark && 66.50 & 84.65 & 51.47 \\
2 & \checkmark   & \checkmark &         && 66.15 & 84.15 & 51.54 \\
3 & \checkmark   & \checkmark & \checkmark && \textbf{68.40} & \textbf{84.95} & \textbf{51.73} \\
\bottomrule
\end{tabular}%
}
\end{table}

\vspace{-1em}
\section{Conclusion}
We present CONQUER, a two-stage framework for Text-Based Person Search. In the training phase, our CARE module enhances cross-modal representation through multi-granularity encoding, complementary pair mining, and context-guided Optimal Transport. At inference, the plug-and-play IQE module adaptively refines ambiguous or incomplete queries by anchor selection and attribute enrichment, all without retraining. Extensive experiments on three public benchmarks demonstrate that CONQUER consistently surpasses strong baselines in both in-domain and cross-domain scenarios. Ablation studies further verify the complementary effects of CARE and IQE. For future work, we aim to further reduce inference latency, improve attribute extraction reliability, and extend CONQUER to broader cross-modal retrieval tasks.

\bibliographystyle{IEEEbib}
\bibliography{strings,refs}

\end{document}